\documentclass{article}
\usepackage{spconf,amsmath,epsfig}
\usepackage{booktabs} 
\usepackage{threeparttable}
\usepackage{epstopdf}
\usepackage{graphicx}
\usepackage{multicol}
\usepackage{multirow}
\usepackage[tight,footnotesize]{subfigure}
\usepackage{algorithm}
\usepackage{algpseudocode}
\usepackage{amsmath}
\usepackage{graphicx}
\usepackage{subfigure}
\usepackage{cite}
\usepackage{balance}

\begin{document}
\title{Mutual Linear Regression-based Discrete Hashing }
\name{Xingbo Liu$^{1}$, Xiushan Nie$^{2*}$, Yilong Yin$^{3}$   \thanks{* corresponding author}}
\address{$^{1}$School of Computer Science and Technology, Shandong University, Jinan, P.R. China\\
$^{2}$School of Computer Science and Technology, Shandong University of Finance and Economics,\\ Jinan, P.R. China\\
$^{3}$School of Software, Shandong University, Jinan, P.R. China\\
sclxb@mail.sdu.edu.cn, niexsh@sdufe.edu.cn, ylyin@sdu.edu.cn 
}
\maketitle

\begin{abstract}
 Label information is widely used in hashing methods because of its effectiveness of improving the precision. The existing hashing methods always use two different projections to represent the mutual regression between hash codes and class labels. In contrast to the existing methods, we propose a novel learning-based hashing method termed stable supervised discrete hashing with mutual linear regression (S2DHMLR) in this study, where only one stable projection is used to describe the linear correlation between hash codes and corresponding labels. To the best of our knowledge, this strategy has not been used for hashing previously. In addition, we further use a boosting strategy to improve the final performance of the proposed method without adding extra constraints and with little extra expenditure in terms of time and space. Extensive experiments conducted on three image benchmarks demonstrate the superior performance of the proposed method.
\end{abstract}


\section{Introduction}
The approximate nearest neighbor search (ANN), which takes a query sample and finds its ANNs within a large database, is vital in many applications. Hashing provides high efficiency in both storage cost and query speed and has become a primary technique in ANN search.

Hashing used in retrieval attempts to encode media data into a string of complex binary codes that preserve the similarity relationships of the original data. Distance in binary codes is calculated using the Hamming distance, which can be performed using hardware with bit-wise XOR operations and provides highly efficient computation compared with other distance calculations \cite{wang2018survey}. Existing hashing methods can be roughly divided into two main categories: data independent and data dependent methods. 
Data independent methods, such as locality-sensitive hashing (LSH) \cite{indyk1998approximate} and its extensions, generate hash codes of the original data with random projections. Data dependent methods (\emph{a.k.a.} learn to hash or learning-based hashing) aim at generating short hash codes by learning the projections under the guidance of the original data. Learning-based hashing is one of the most accurate hashing methods because it can provide better retrieval performance by analyzing the underlying characteristics of the data. The existing learning-based hashing methods can be roughly divided into two main categories:
unsupervised methods 
\cite{gong2013iterative} and supervised methods \cite{shen2015supervised}    \cite{gui2016supervised}.
Unsupervised hashing does not use label information for the training samples. 
In contrast, supervised hashing methods make full use of class labels. 
Deep supervised hashing is proposed recently, which uses deep learning to perform feature learning for hashing \cite{shen2017deep}. In general, deep supervised hashing can perform significantly better than non-deep supervised hashing.
 
Discrete constraints are an important factor in learning-based hashing, which usually give rise to mixed integer optimization problems (usually NP-hard). A relaxation strategy is adopted to address this issue, which requires discarding the discrete constraints in the optimization procedure and then transforming real values into hash codes using thresholding~\cite{ Wang2015Hamming}. However, these relaxed methods usually suffer from accumulated quantization error and local optima~\cite{gui2018fast}. To tackle this problem, Shen et al. propose a novel method named supervised discrete hashing (SDH) ~\cite{shen2015supervised}, which can produce discrete hash codes directly and without relaxation. However, SDH is time-consuming and less stable on some level. To solve this problem, Gui et al. develop a method named fast supervised discrete hashing (FSDH)~\cite{gui2018fast}, which can stabilize the generation of hash codes and speed up the training process. However, FSDH still seems to be unstable and suffers from local optima when used to bridge the semantic gap between a discrete hash code and discrete label matrix using one simple projection. 

In order to address the aforementioned issues, we propose a novel method termed stable supervised discrete hashing with mutual linear regression (S2DHMLR). In contrast to previous hashing methods, we propose a novel utilization of label information with mutual linear regression. Specifically, we regress the hash codes to the corresponding label matrix with a linear projection and regress the label matrix to the corresponding hash codes using the same linear projection simultaneously. The learned linear projection can describe a stable and unique correlation between a hash code and a label matrix. To the best of our knowledge, this strategy has not been not used for hashing previously. The main contributions of this study are summarized as follows:
\begin{itemize}
\item Only one projection is used to describe the mutual regression between hash codes and class labels, which makes the hashing method more stable and precise.

\item We propose a hash boosting strategy that boosts the performance of the proposed method, where more efficient parameters can be learned using the boosting strategy. 

\item Experiments based on three large-scale datasets show that
the proposed method can provide superior performance under various scenarios.
\end{itemize}

\begin{figure}[tp]
\centering\includegraphics[width=0.47\textwidth]{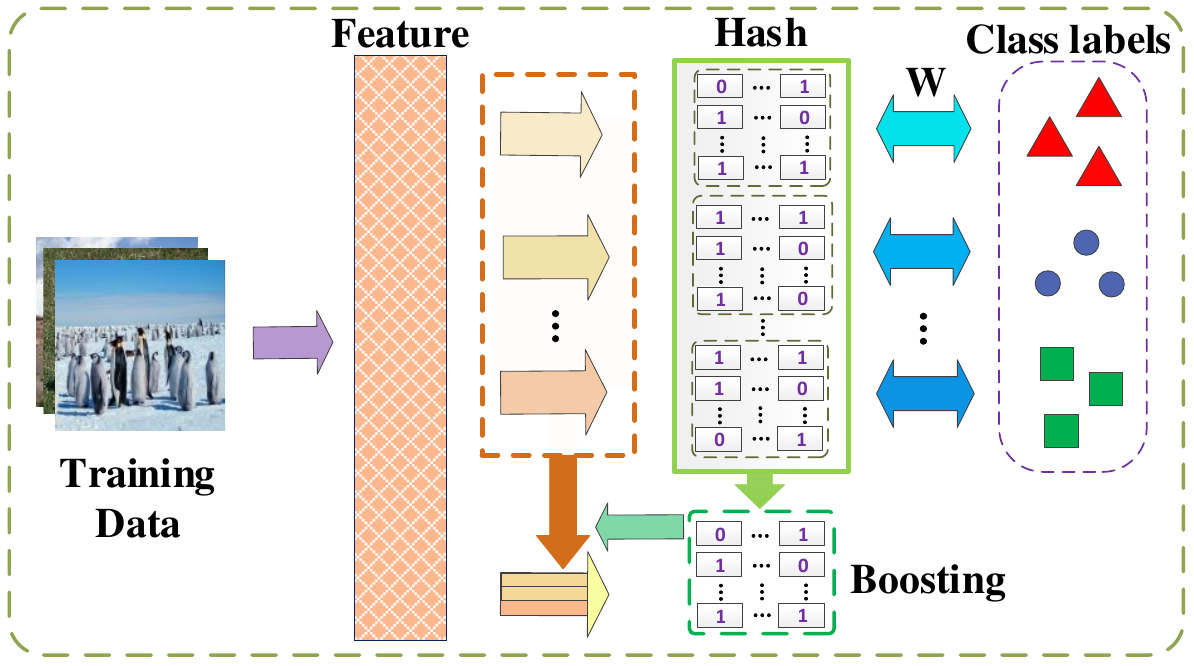}
\caption{Framework of the proposed S2DHMLR method. We can not only run the proposed method once to obtain a stable solution but also  run the method several times to obtain a more precise and stable solution by adopting the hash boosting strategy.}
\end{figure}
 
\section{Proposed Method}
The S2DHMLR framework is illustrated in Fig. 1, where the method involves two regression steps. The first step involves regressing the original feature to hash code, and the other step involves regression between hash codes and class labels. In contrast to the existing supervised hashing methods, which always regress the hash codes to their corresponding labels (e.g., ~\cite{shen2015supervised}), or regress class labels to hash codes using a different projection (such as ~\cite{gui2018fast}), both mutual regression steps between hash codes and labels are used with the same projection in S2DHMLR. The use of a single projection makes the proposed method more stable and precise. 

Moreover, we propose a boosting strategy in the S2DHMLR to learn a more optimal projection between the original feature and hash codes. 

\subsection{Formulation}
Assume that we have a training set $\bf{V}$ consisting of $n$ instances, i.e., ${\bf{V}} = \left ( {\bf{v}}_i\right ) _{i = 1}^n$. Each instance can be represented by a $d$-dimensional feature. Moreover, a semantic label matrix ${\bf{Y}} = \{ {{\bf{y}}_i}\} _{i = 1}^n$ is also available with ${{\bf{y}}_i= \{ {{{y}}_{ij}}} \} \in R^{c \time 1}$ being the label vector of the $i_{th}$ instance, where $c$ is the number of categories in the training set. If the $i_{th}$ instance belongs to the $j_{th}$ semantic category, ${y_{ij}} = 1$, and $-1$ otherwise. The hash matrix is defined as ${\bf{H}}=\{{\bf{h}}_{i}\}_{i = 1}^n$. Moreover, ${\rm{||}}{\bf{W}}||$ and ${{\bf{W}}^T}$ denote the $\ell_{2}$-norm and transpose of a matrix $\bf{W}$, respectively.

Given the hash matrix ${\bf{H}}$ and label matrix ${\bf{Y}}$, a linear model is  commonly used to describe the correlation due to its efficiency. Typically, SDH ~\cite{shen2015supervised} attempts to find a projection from a hash matrix ${\bf{H}}$ to a label matrix ${\bf{Y}}$. It can be formulated as
\begin{equation}
\min\limits_{{\bf {P_{H}}},{\bf {H}}}\left \| {\bf {Y}}-{\bf{P_{H}}}^{T}\bf{H} \right \|^{2}+\lambda \left\|{\bf{P_{H}}}\right\|^{2}
  \quad \textup{s.t.}  \quad {\bf {H}} \in \left  \{ -1,1 \right \}^{L\times n},
\end{equation}
where $\lambda$ is a regularization parameter.
The closed-form solution of ${\bf{P_{H}}}$ is
\begin{equation}
{\bf{P_{H}}=({\bf{HH}}^{T}}+\lambda {\bf {I}})^{-1}{\bf{H}}{\bf{Y}}^{T}.
\end{equation}

However this strategy is time-consuming and less stable on some level. To tackle this problem,
FSDH \cite{gui2018fast} attempts to find a projection between a label matrix ${\bf{Y}}$ and a hash matrix ${\bf{H}}$. It can be formulated as follows:
\begin{equation}
\min\limits_{{\bf {P_{Y}}},{\bf {H}}}\left \| {\bf {H}}-{\bf{P_{Y}}}{\bf{Y}} \right \|^{2}+\lambda \left\|{\bf{P_{Y}}}\right\|^{2}
  \quad \textup{s.t.}  \quad {\bf {H}} \in \left \{ -1,1 \right \}^{L\times n},
\end{equation}
The closed-form solution of ${\bf{P}}$ is
 \begin{equation}
{\bf{P_{Y}}}={\bf{H}}{\bf{Y}}^{T}({\bf{YY}}^{T}+\lambda {\bf {I}})^{-1}.
 \end{equation}
Generally, ${\bf P_{H}}\neq {\bf P_{Y}}$. A brief proof is shown below.

Assume that ${\bf P_{H}}= {\bf P_{Y}}$; according to Eq. (2) and Eq. (4), we have
\begin{equation}
({\bf{HH}}^{T}+\lambda {\bf {I}})^{-1}{\bf{H}}{\bf{Y}}^{T}={\bf{H}}{\bf{Y}}^{T}({\bf{YY}}^{T}+\lambda {\bf {I}})^{-1}.
\end{equation}
Therefore,
\begin{equation}
{\bf{H}}{\bf{Y}}^{T}({\bf{YY}}^{T}+\lambda {\bf {I}})=({\bf{HH}}^{T}+\lambda {\bf {I}}){\bf{H}}{\bf{Y}}^{T}.
\end{equation}
Subsequently, this yields the following equation:
\begin{equation}
{\bf{Y}}^{T}{\bf{Y}}={\bf{H}}^{T}{\bf{H}}.
\end{equation}
Obviously, Eq. (7) does not hold because the label matrix is not equivalent to the hash matrix. Therefore, the assumption is false and ${\bf P_{H}}\neq {\bf P_{Y}}$.
In this study, we try to find one projection to replace ${\bf P_{H}}$ and ${\bf P_{Y}}$, and obtain more stable and precise performance for retrieval task.

In general, existing methods either regress the hash code to the class label or vice versa. If we consider the hash code as a kind of sample representation, and the class label as the representation of semantic latent space, then mutual regression between hash codes and class labels can be formulated as a linear auto-encoder. In contrast to existing methods, S2DHMLR uses the same projection for the encoder and decoder. In other words, the projection matrix ${\bf{W}}$ learns to map the label matrix ${\bf{Y}}$ to the hash matrix ${\bf{H}}$, and the transpose of ${\bf{W}}$ (i.e. ${\bf{W}}^{T}$) is used to map the hash matrix ${\bf{H}}$ to the label matrix ${\bf{Y}}$. This process can be written as
\begin{equation}
\begin{split}
&\min\limits_{{\bf {W}},{\bf {H}}}\left \| {\bf {Y}}-{\bf{W}}^{T}\bf{H} \right \|^{2}+\alpha \left \| {\bf {H}}-{\bf{W}}\bf{Y} \right\|^{2}+\lambda \left\|{\bf{W}}\right\|^{2} \\
& \textup{s.t.}  \quad {\bf {H}} \in \left \{ -1,1 \right \}^{L\times n},
\end{split}
\end{equation}
where $\alpha$ is a parameter that represents the trade-off between these terms.
It is noteworthy that this strategy is different from previous methods based on matrix factorization 
~\cite{Ding2014Collective}. The projection ${\bf {W}}$ is a strong correlation between the label matrix and the hash code. ${\bf{W}}$ and ${\bf{W}}^{T}$ can be seen as each other's inverse mapping. One can prove that inverse mapping is unique using set theory \cite{jech2013set}. Thus, the optimal solution to ${\bf {W}}$ seems to be unique and stable. As a result, the hash method will learn well for out-of-samples according to Bousquet and Elisseeff’s theory \cite{Lemma2002Stability}.

In addition, assume $\bf{P}$ is a projection between an original feature and hash code. Regression from the original feature to the hash code can be formulated as follows:
 \begin{equation}
\min\limits_{{\bf{P}}}\left\|{\bf{H}}-{\bf{P}}^{T}{\bf{V}} \right\|^{2}+\lambda \left\|{\bf{P}}\right\|^{2}\quad
{s.t.}\quad {\bf{H}}\in \left \{ -1,1 \right \}^{L\times n}.
\end{equation}

Finally, the formulation of S2DHMLR is 
\begin{equation}
\begin{split}
&\min\limits_{{\bf {W}},{\bf {H}},{\bf {P}}}\left \| {\bf {Y}}-{\bf{W}}^{T}\bf{H} \right \|^{2}+\alpha \left \| {\bf {H}}-{\bf{W}}\bf{Y} \right\|^{2} \\
&+\beta\left \|{\bf {H}}-{\bf{P}}^{T} \bf{V} \right \|^{2}+\lambda (\left \| \bf P \right \|^{2}+\left\|{\bf{W}}\right\|^{2})\\
& \textup{s.t.}  \quad {\bf {H}} \in \left \{ -1,1 \right \}^{L\times n},
\end{split}
\end{equation}
where $\alpha$ and $\beta$ are parameters.

\subsection{Optimization}
It is challenging to optimize Eq. (10) directly as it is non-convex and non-continuous. However, this non-differentiable problem can be solved with an iterative framework using the following steps until convergence.

Step 1: Learn the mutual projection $\bf{W}$ with the other variables fixed. The problem in Eq. (10) becomes
\begin{equation}
\min\limits_{{\bf {W}}} \left \| {\bf {Y}}-{\bf{W}}^{T}\bf{H} \right \|^{2}+\alpha \left \| {\bf {H}}-{\bf{W}}\bf{Y} \right\|^{2}+ \lambda \left\|{\bf{W}}\right\|^{2}
  \end{equation}
Setting the derivative of Eq. (11) with respect to $\bf{W}$ to zero yields
\begin{equation}
{\bf{H}}{\bf{H}}^{T}{\bf{W}}+\alpha {\bf{W}}({\bf{Y}}{\bf{Y}}^{T}+\lambda {\bf{I}}) =(1+\alpha){\bf{H}}{\bf{Y}}^{T}.
\end{equation}
which is a standard Sylvester equation \cite{Golub2009Matrix} and can be solved analytically in closed form.

Step 2: Learn the binary code ${\bf{H}}$ with the other variables fixed. The problem in Eq. (10) becomes
\begin{equation}
\begin{split}
&\min\limits_{{\bf {H}}} \left \| {\bf {Y}}-{\bf{W}}^{T}\bf{H} \right \|^{2}+\alpha \left \| {\bf {H}}-{\bf{W}}\bf{Y} \right\|^{2} +\beta\left \|{\bf {H}}-{\bf{P}}^{T} \bf{V} \right \|^{2}\\ &\textup{s.t.} \quad {\bf{H}}\in \left \{ -1,1 \right \}^{L\times n}.
\end{split}
\end{equation} 

Eq. (13) can be reformulated as
\begin{equation}
\begin{split}
&\min\limits_{{\bf {H}}} \left \| {\bf{Y}} \right \|^{2}-2Tr( {\bf{Y}}^{T}{\bf{W}}^{T}{\bf{H}})+\left \| {\bf{W}}^{T}{\bf{H}} \right \|^{2}\\
&+\alpha(\left \| {\bf{H}} \right \|^{2}-2Tr({\bf {H}}^{T}{\bf{W}}{\bf{Y}}+\left \| {\bf{W}}{\bf{Y}} \right \|^{2})\\
&+\beta (\left \| {\bf{H}} \right \|^{2}-2Tr({\bf {H}}^{T}{\bf{P}}^{T}{\bf{V}})+\left \| {\bf{P}}^{T}{\bf{V}}\right \|^{2})\\
&\textup{s.t.} \quad {\bf{H}}\in \left \{ -1,1 \right \}^{L\times n}.
\end{split}
\end{equation}
where $Tr(\cdot)$ is the trace. Since $\left \| {\bf{H}} \right \|^{2}=L\ast n$, Eq. (14) can be rewritten as
\begin{equation}
\begin{split}
&\min\limits_{{\bf {H}}} \left \| {\bf{W}}^{T}\bf{H} \right \|^{2}+ Tr({\bf {H}}^{T}{\bf{M}})\\
&\textup{s.t.} \quad {\bf{H}}\in \left \{ -1,1 \right \}^{L\times n}.
\end{split}
\end{equation}
where ${\bf{M}} =(1+\alpha) {\bf{W}}{\bf{Y}} +\beta {\bf{P}}^{T}{\bf{V}}$. 

Although Eq. (15) is difficult to solve as ${\bf{H}}$ is discrete, we can directly leverage the
discrete cyclic coordinate descent (DCC) approach ~\cite{shen2015supervised} to learn ${\bf{H}}$ bit-by-bit iteratively. 
Specifically, define ${\bf{h}}^{T}$ as the $l_{th}$ row of matrix ${\bf{H}}$, $l=1,\cdots, L$ and  ${\bf{H}}^{'}$ as the matrix ${\bf{H}}$ excluding $\bf{h}$. Analogously, define  ${\bf{q}}^{T}$ as the $l_{th}$ row of matrix ${\bf{W}}$ and  ${\bf{W}}^{'}$ as the matrix ${\bf{W}}$ excluding ${\bf{q}}$. Next, define ${\bf{w}}^{T}$ as the $l_{th}$ row of matrix ${\bf{W}}$. The analytic solution of ${\bf{h}}$ can be written as:
\begin{equation}
    {\bf{h}}=sgn({\bf{m}}-{{\bf{H}}'}^{T}{\bf{W}}'\bf{q}),
\end{equation}
where $sgn(\cdot)$ is a sign function.

Step 3: Learn the projection matrix ${\bf{P}}$ while holding the other variables fixed. The problem in Eq. (10) becomes:
 \begin{equation}
\min\limits_{{\bf{P}}}\left\|{\bf{H}}-{\bf{P}}^{T}{\bf{V}} \right\|^{2}+\lambda \left\|{\bf{P}}\right\|^{2}.
\end{equation}
The closed-form solution of ${\bf{P}}$ is 
 \begin{equation}
{\bf{P}}=({\bf{VV}}^{T}+\lambda {\bf{I}})^{-1}{\bf{V}}{\bf{H}}^{T}.
 \end{equation}
 
In short, Eq. (10) can be solved using cyclic coordinate descent. In each step, the optimal solution is obtained in the sub-problem. The convergence of this strategy is proven using optimization theory ~\cite{Tseng2001Convergence} ~\cite{Xu2015A}. 

\subsection{Hash Boosting}

We can learn a projection $\bf{P}$ for the out-of-sample extension from the optimization problem in Eq. (10). However, in this study, we further propose a hash boosting strategy to learn a better projection ${{\bf{P}}_F}$.

It is known that bit balance, meaning each bit has an approximately 50\% chance
of being +1 or $- 1$, can avoid trivial solutions to the optimization problem, thus making the hash code more efficient~\cite{wang2018survey}. Given a hash matrix, the $i_{th}$ row of this matrix represents the $i_{th}$ bit dimension of all training samples; we denote the absolute value of the sum of the $i_{th}$ row in the hash
matrix as the balance degree. Obviously, the bit balance of this row is much better when the balance degree is approximately zero. 

According to bit balance, two steps are included in the proposed hashing boosting strategy. In the first step, given the hash code of length $L$, we first run the proposed method S2DHMLR $T$ times to obtain $T$ different hash matrices as training samples, whose size is $L \times n$ ($n$ is total number of samples). We subsequently concatenate the $T$ hash matrices in the column direction to construct a concatenated matrix. Finally, we select the first $L$
rows with minimum balance degrees from the concatenated matrix to
construct the final hash matrix ${{\bf{H}}_F}$ of training samples. To demonstrate the construction of the hash matrix ${{\bf{H}}_F}$, we present an example in Fig. 2, where  $L=3$, $T=3$, and $n=4$. According to Eq. (18), we can see that the row of the projection matrix ${{\bf{P}}}$ corresponds to the bit row of the hash matrix ${{\bf{H}}}$. Therefore, in the second step,  we select some rows in the $T$ learned projections (${\bf{P}}_1$, ${\bf{P}}_2$, $\cdots$, ${\bf{P}}_T$) corresponding to each row of  ${{\bf{H}}_F}$ to construct the final projection matrix ${{\bf{P}}_F}$.
With the new projection ${{\bf{P}}_F}$, we can obtain more stable and precise hash code for the out-of-samples. 
\begin{figure}[tp]
\centering\includegraphics[width=0.45\textwidth]{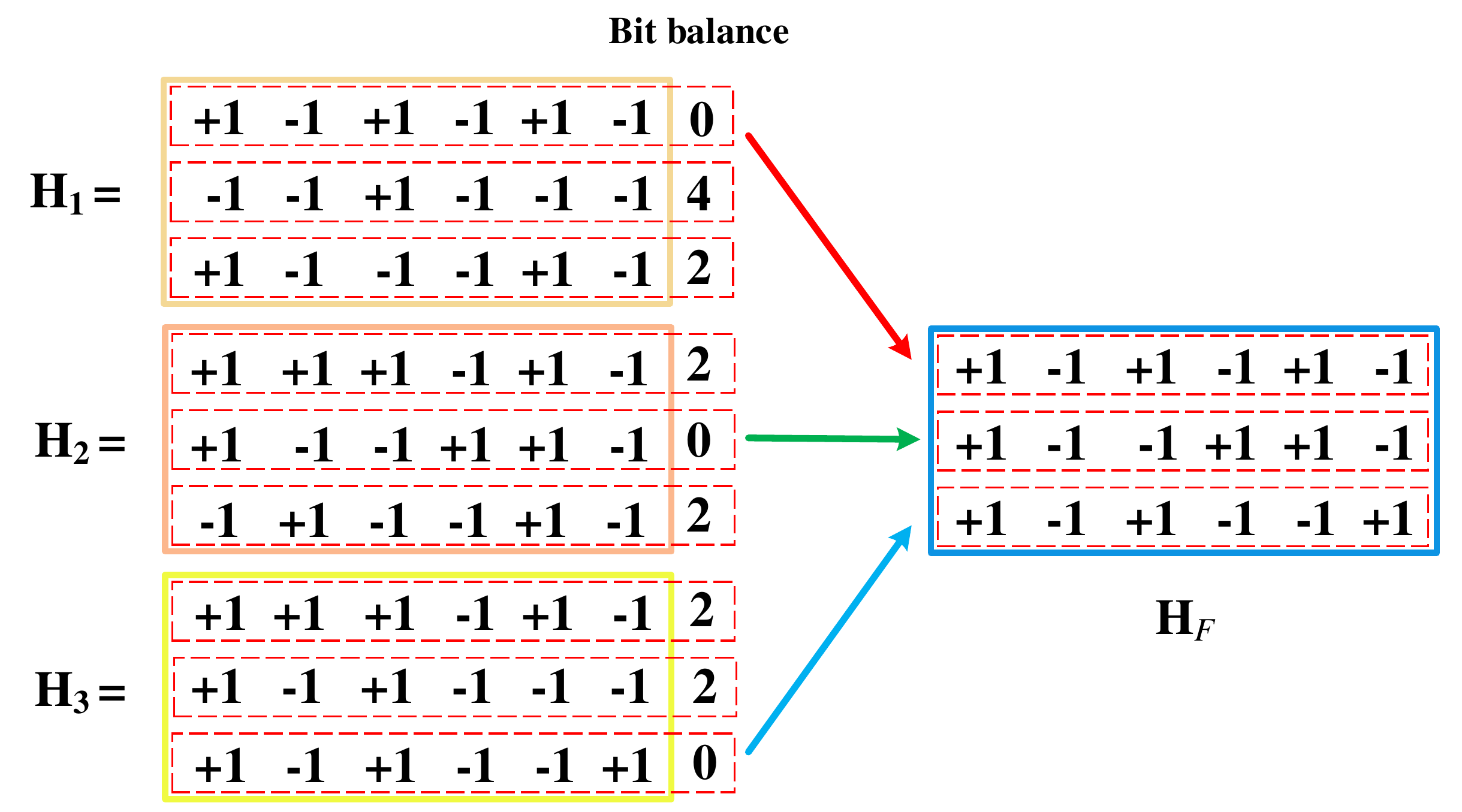}
\caption{Illustration showing construction of the hash matrix ${{\bf{H}}_F}$.}
\end{figure}

\subsection{Time Complexity}
The time complexity for learning the projection $\bf{P}$ and hash code  $\bf{H}$ are $O(2nd^{2}+ndL)$ and $O(ncL^{2}+ndL)$, respectively, while the time complexity for learning the projection $\bf{W}$ is $O(nc^{2}+nL^{2}+ncL+L^{3})$. Therefore, the total training time complexity of the proposed method is $O(ncL^{2}+ndL+L^{3})$. With the boost strategy, the time complexity is $T \cdot O(ncL^{2}+ndL+L^{3})+O(TLlgL+NTL)$. However, $T$ is always small because we find
that the precision increases very slowly when $T$ is larger than a small value. We set $T$ to 3 in the experiments.

\section{Experiments}

\subsection{Experimental Settings}
\begin{table*}[htp]
  \centering
  \fontsize{7}{9}\selectfont
  \begin{threeparttable}
  \caption{Performance in terms of mAP score.}
  \begin{tabular}{c|c|c|c|c|c|c|c|c|c|c|c|c}
    \toprule
    \multirow {2}{*}{Method}&\multicolumn{4}{|c}{CIFAR-10} &\multicolumn{4}{|c}{MS-COCO} &\multicolumn{4}{|c}{NUS-WIDE} \cr
    \cmidrule(lr){2-5} \cmidrule(lr){6-9} \cmidrule(lr){10-13}
    \!&12 bits\!&24 bits\!&\!32 bits\!&\!48 bits\!&\!12 bits\!&\!24 bits\!&\!32 bits\!&\!48 bits\!&12 bits\!&24 bits\!&\!32 bits\!&\!48 bits\!\cr
    \midrule
    
    SH &0.2704&0.2908&0.2898 &0.2961&0.6490& 0.6501 &0.6529&0.6800&0.5958 &0.5986&	0.6058 & 0.6070\cr

    PCA-ITQ &0.2767&0.3554&0.3398&0.3562 &0.5328&0.6281&0.6578&	0.6907&	0.3181&	0.4050&	0.4544   &	0.6016\cr

    PCA-RR &0.2903&	0.3054	&0.2938	&0.3172 &0.5590&0.5677&	0.6481&	0.6594&	0.5519&0.5649&0.5500	&0.6282\cr

    MFH &0.2991&0.3345&0.3473&0.3623&0.6171 &0.6330 &0.6470 &0.6500 & 0.5820 & 0.6088&  0.6244& 0.6315 \cr

    SDH & 0.5111& 0.6358&0.6507&0.6626 &0.5482&0.6037&0.6489&0.6531&0.4978&0.5022 & 0.5775 &0.7350\cr

    COSDISH &0.3820&0.4366&0.4854 &0.5295 &0.5230&0.5348&0.5390& 0.6164&0.3099	&0.3135&  0.4242&	0.4274\cr

    FSDH &0.5370&0.6218&0.6526&	0.6632 &0.5898&	0.7162&	0.7197&	0.7235&	0.6845&	0.7241&	0.7574&	0.7763\cr
    \hline
   DHN & 0.6805& 0.7213& 0.7233 &0.7332 &0.7440 &0.7656 &0.7691 &0.7740 &0.7719 &0.8013 &0.8051 &0.8146\cr

   DSH & 0.6441 &0.7421 &0.7703& 0.7992 &0.6962 &0.7176& 0.7156 &0.7220 &0.7125 &0.7313& 0.7401& 0.7485\cr
 \hline 
   S2DHMLR &{{0.5432 }}&{{0.6501}}&{ {0.6606}}&{{0.6818}}&{{0.6751}}&{{0.7414 }}&{{0.7890    }}&{{0.8141}}&{0.7173}&{{0.7768}}&{{0.7810}}&{{0.7852}}\cr 
 
   
  S2D-boost &{{0.6545 }}&{{0.6906}}&{ {0.7001}}&{{0.7020}}&{{0.7957}}&{{0.7878 }}&{{0.8337    }}&{{0.8475}}&{0.7513}&{{0.7822}}&{{0.7976}}&{{0.7977}}\cr
   
    \bottomrule
    \end{tabular}
   \end{threeparttable}
\end{table*}

\begin{figure*}[htb]
\centering 
\subfigure[Based on CIFAR-10]{
\includegraphics[width=0.25\textwidth]{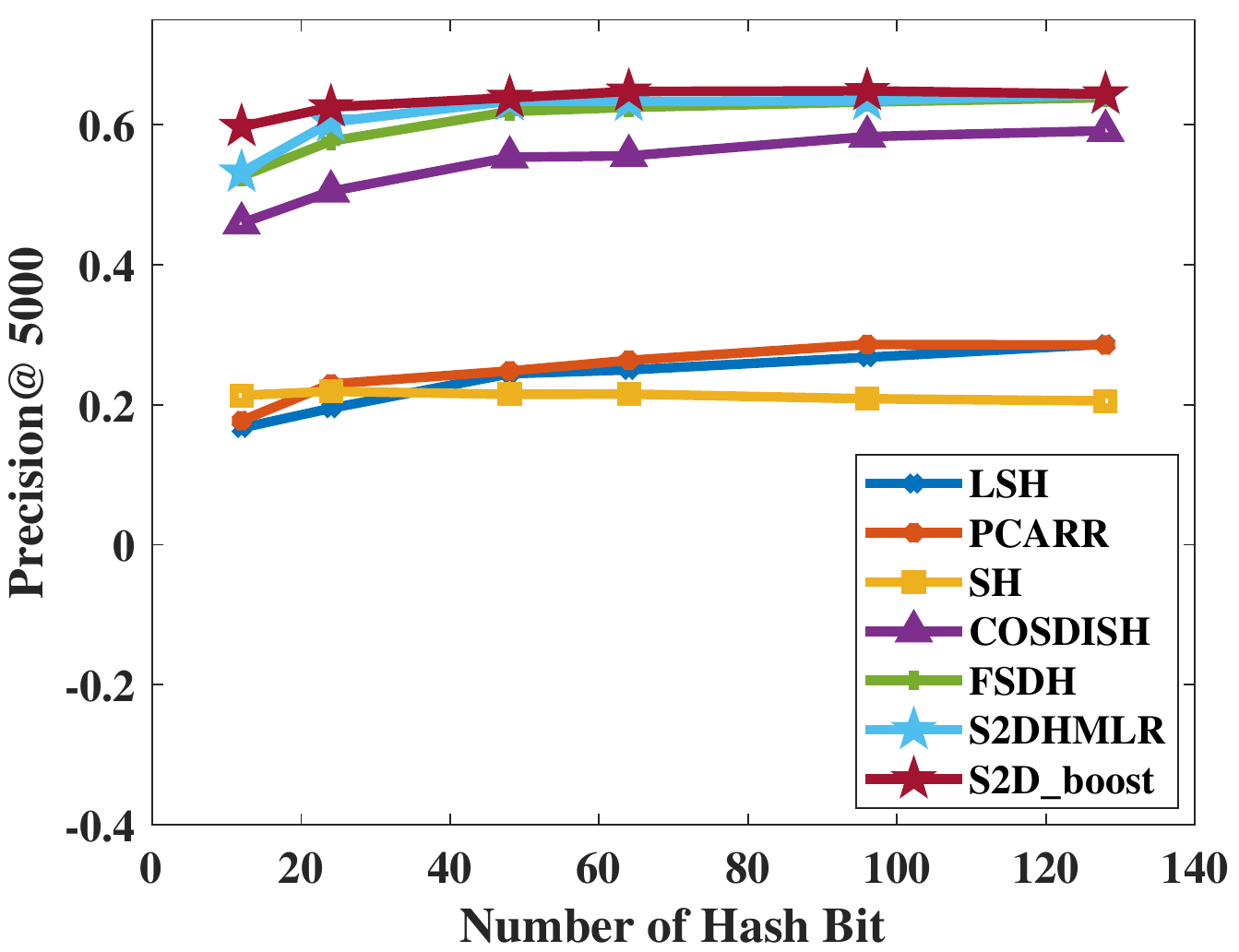}}
\subfigure[Based on MS-COCO]{
\includegraphics[width=0.25\textwidth]{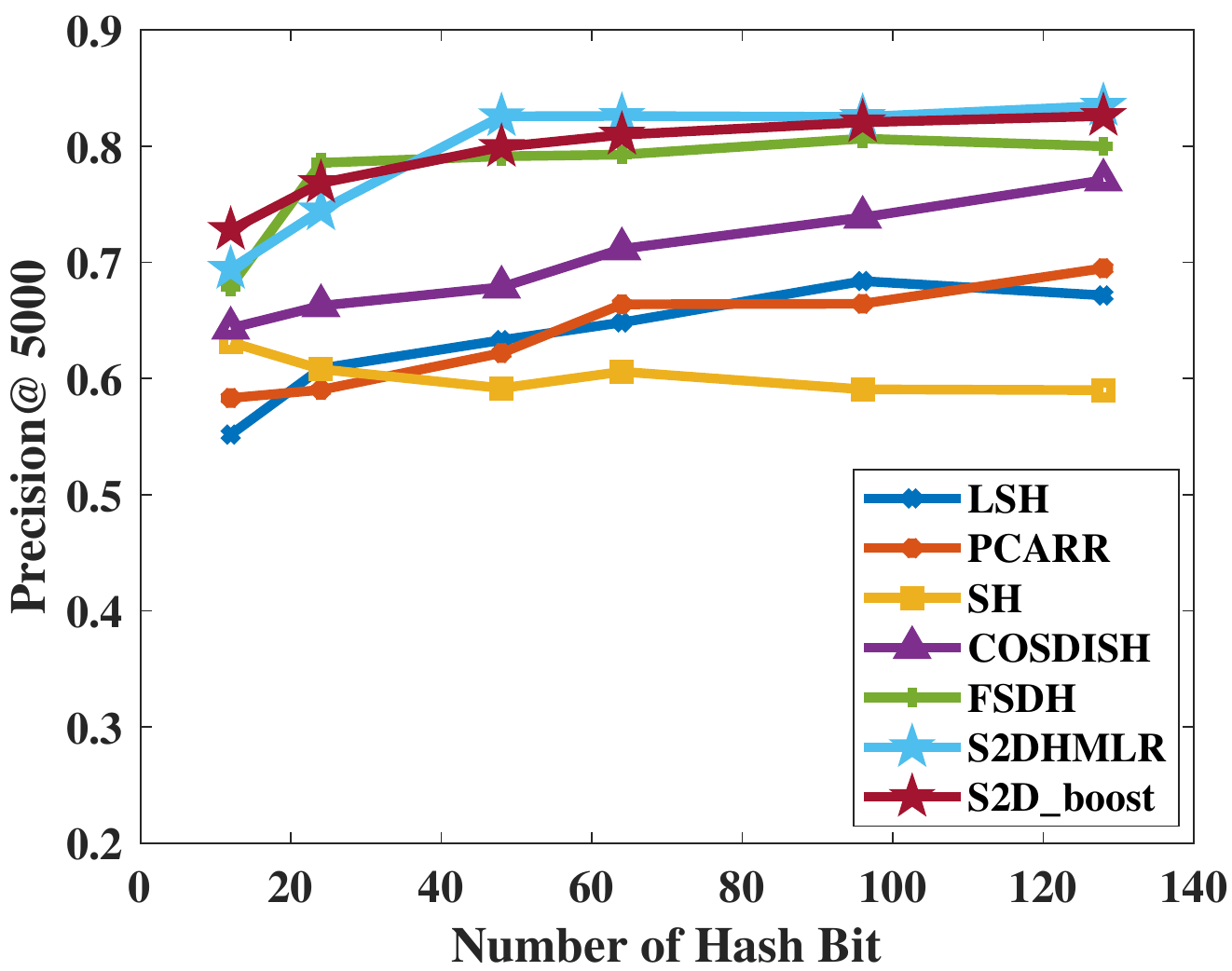}}
\subfigure[Based on NUS-WIDE]{
\includegraphics[width=0.25\textwidth]{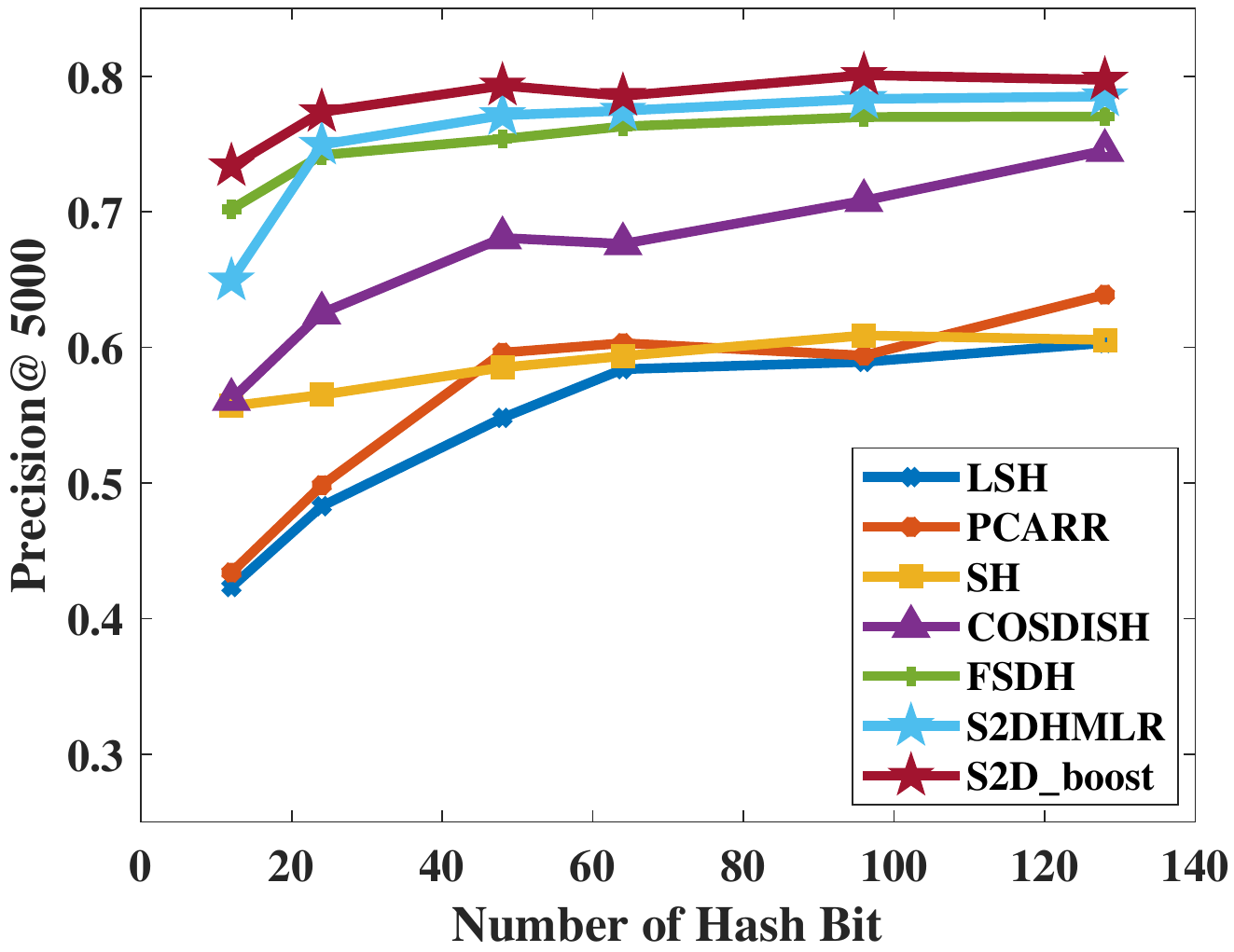}}
\caption{Performance in terms of  precision score based on three benchmark datasets. }
\end{figure*}
\begin{figure*}[htb]
\centering
\subfigure[Based on CIFAR-10]{
\includegraphics[width=0.25\textwidth]{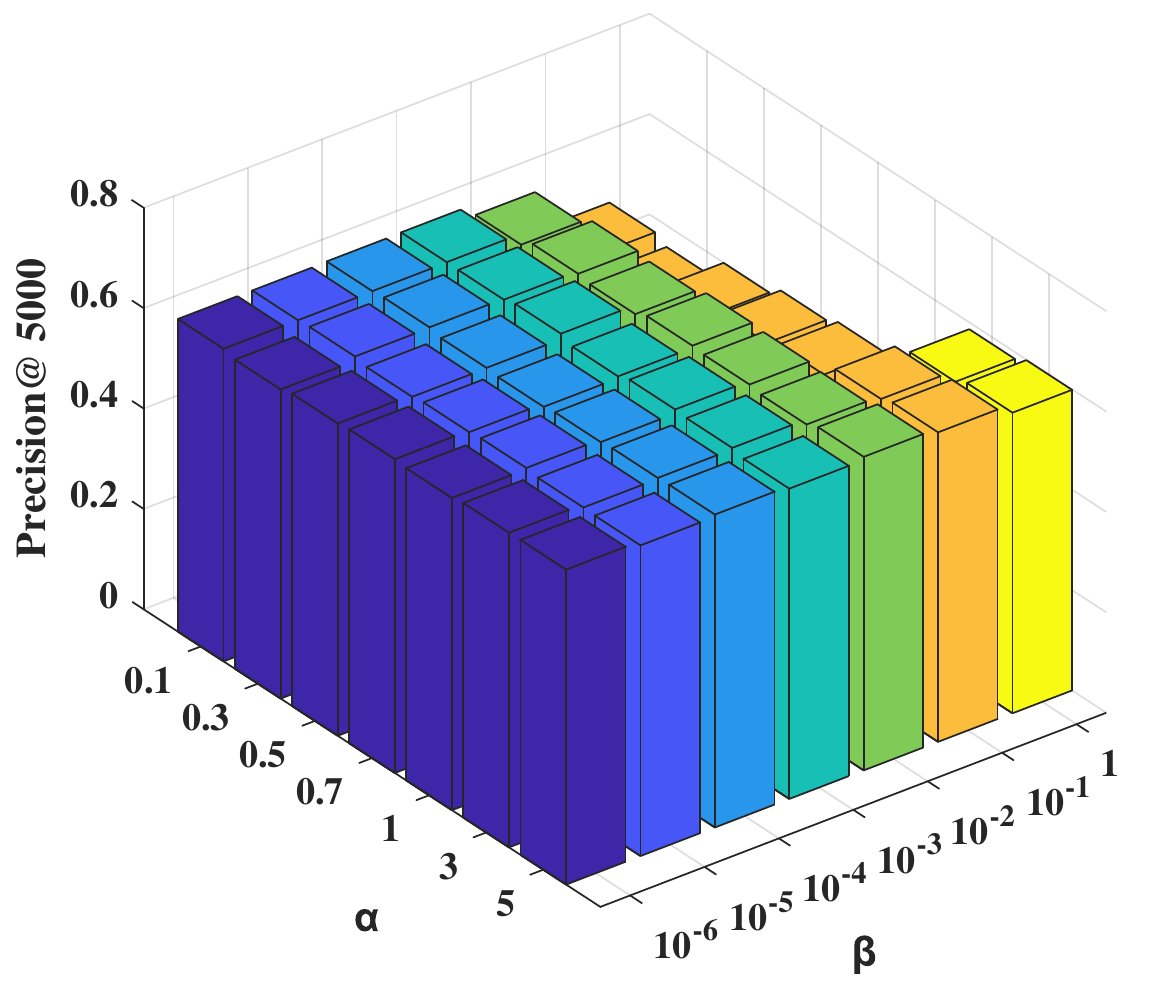}}
\subfigure[Based on MS-COCO]{
\includegraphics[width=0.25\textwidth]{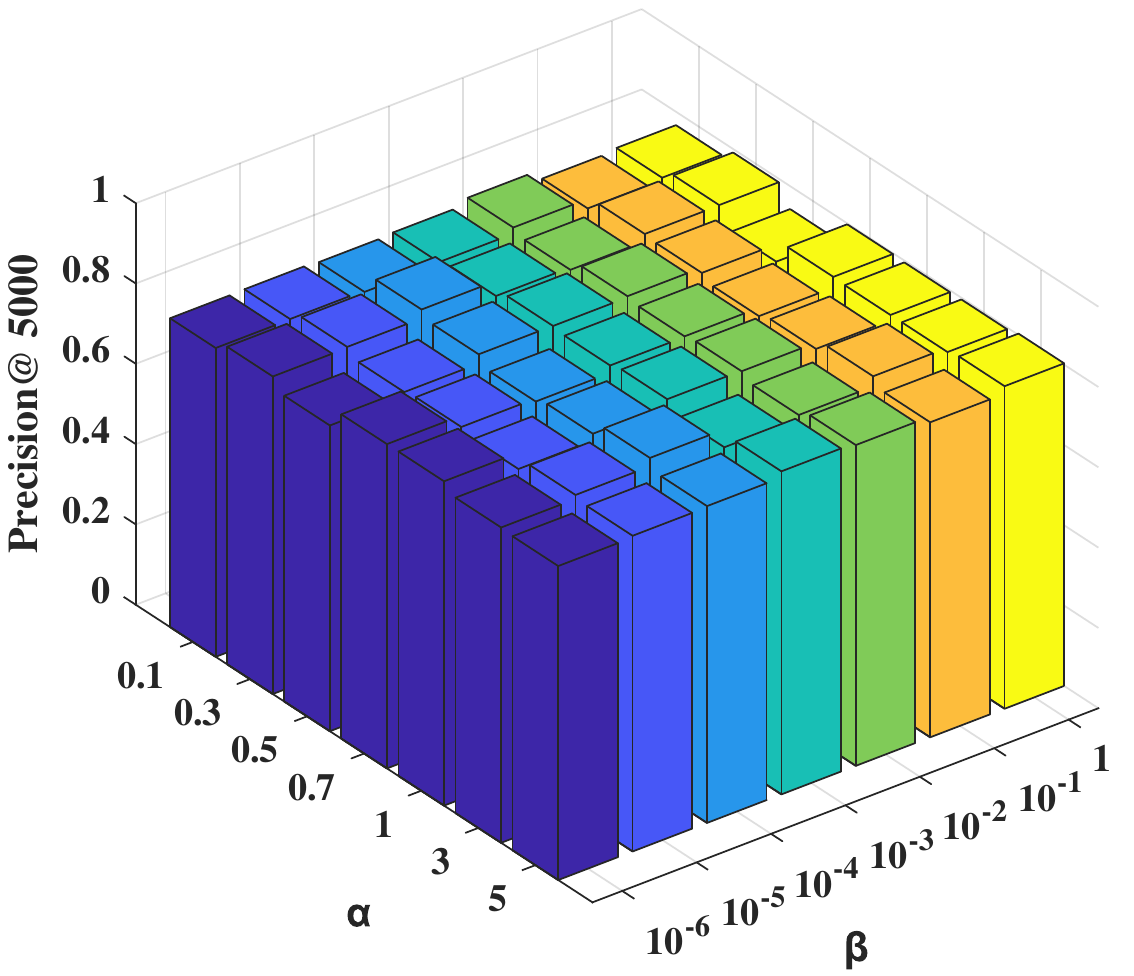}}
\subfigure[Based on NUS-WIDE]{
\includegraphics[width=0.25\textwidth]{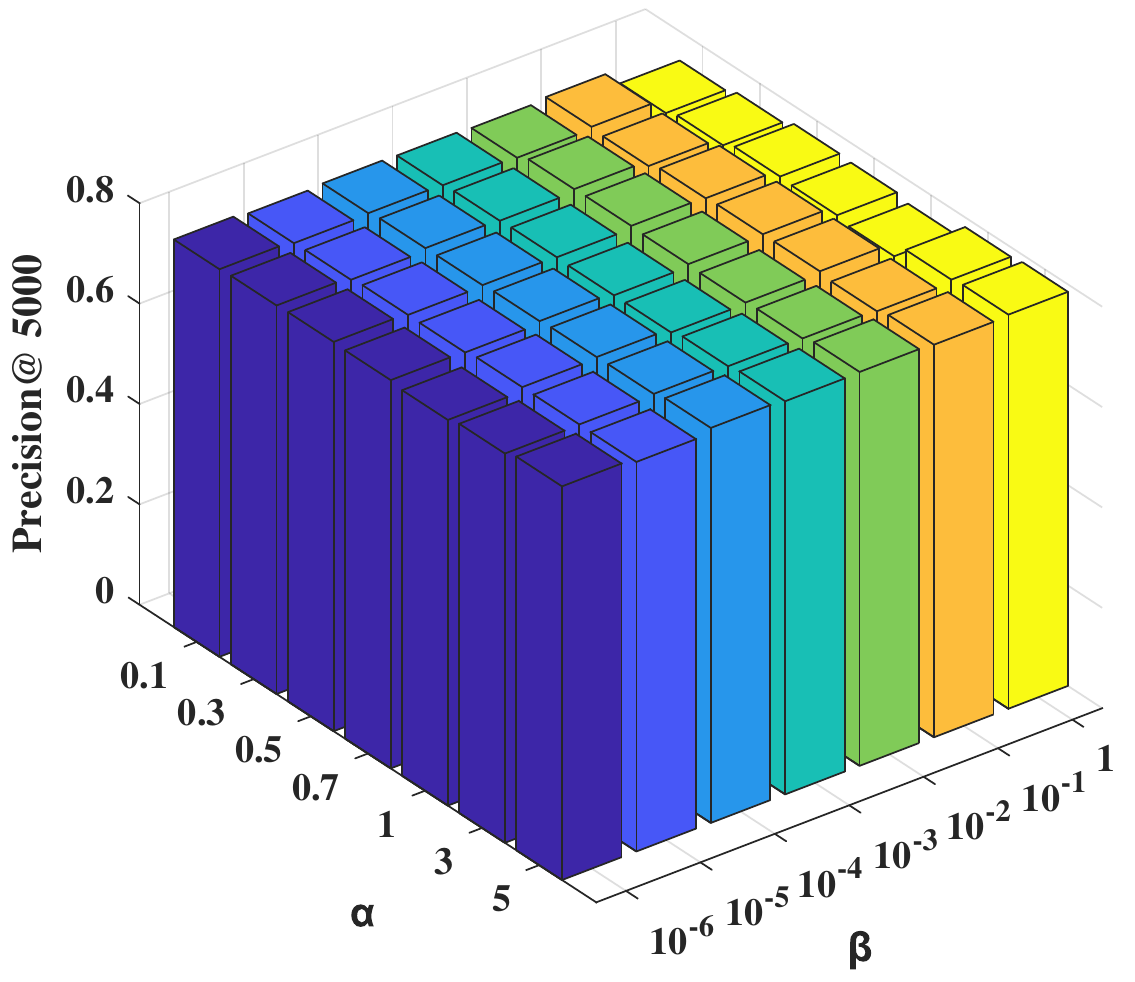}}
\caption{ Precision@ 5000 with different setting of $\alpha$ and $\lambda$ based on three benchmark datasets.}
\end{figure*}

We use three different image datasets in our experiments: CIFAR-10 \cite{krizhevsky2009learning}, MS-COCO \cite{lin2014microsoft}, and NUS-WIDE \cite{chua2009nus}. We use a convolutional neural network (CNN) model called the CNN-F model \cite{chatfield2014return} to perform feature learning. In addition, a radial basis function is used to reduce the number of parameters. Specifically, the 4,096-D deep features extracted by the CNN-F model are mapped to 1,000-D features.
We perform five runs of our method and average their performance for purposes of comparison. Regarding the experimental parameters, we empirically set $\alpha=\lambda = 1$ and $\beta=10^{-5}$. 

To evaluate the proposed method, we use an evaluation metric known as mean average precision (mAP), which is widely used in image retrieval evaluation.
mAP is the mean of the average precision values obtained for the top retrieved samples.

\subsection{Experimental Results and Analysis}
We compare S2DHMLR with the following methods:
spectral hashing (SH) ~\cite{weiss2009spectral},
principle component analysis (PCA)-iterative quantization (PCA-ITQ)  ~\cite{gong2011iterative},
PCA-random rotation (PCA-RR) ~\cite{gong2011iterative},
collective matrix factorization hashing (MFH) ~\cite{Ding2014Collective},
supervised discrete hashing (SDH) ~\cite{shen2015supervised},
column sampling based discrete supervised hashing (COSDISH) ~\cite{kang2016column},
fast supervised discrete hashing (FSDH) ~\cite{gui2018fast}, 
deep hashing network (DHH) ~\cite{zhu2016deep}, and
deep supervised hashing (DSH) ~\cite{Liu2016Deep}.
SH, PCA-ITQ, and PCA-RR are unsupervised hashing methods, while SDH, COSDISH, and FSDH are supervised hashing methods. In addition, DHH and DSH are deep learning-based methods.
All hyper-parameters are initialized as suggested in the original publications. 

Table 1 shows the mAP value for each method with the hash code length ranging from 12 to 48 bits, where S2D-boost represents the S2DHMLR method with the boosting strategy. The mAP performance of S2DHMLR is much higher than that of the other non-deep learning-based methods on the three benchmark datasets. Obviously, the proposed S2DHMLR is not a deep learning model. However, S2DHMLR also shows performance that is comparable to the deep hashing methods, and S2DHMLR exhibits stable improvement when it is boosted with the boosting strategy in all cases.

Figure 2 shows the precision@$5000$ of S2DHMLR method and other methods when the hash codes length ranged from 12 bits to 128 bits. One can see that S2DHMLR and its variant version with boosting exhibits better performance. We can also see that the precision is higher for longer hash codes in most cases, which is reasonable because long hash codes contain more information about the samples. 

In order to verify the stability of the proposed method, we conduct some experiments with different parameter settings. Figure 3 shows the precision@$5000$ from S2DHMLR when ${\alpha}$ ranges from $0.1$ to $5$ and ${\beta}$ ranges from $10^{-6}$ to $1$. The results show that the S2DHMLR method exhibits satisfactory stability and sensitivity.

\section{Conclusion}
In this study, we propose a method called stable supervised discrete hashing with mutual linear regression. Semantic label information is leveraged to learn one mutual projection between a hash code and a label matrix, and we propose a fusion strategy to boost projection learning. The proposed method can produce more stable and precise performance in various scenarios. Experiments conducted on three benchmark datasets indicate that the proposed method exhibits superior performance.

\bibliographystyle{IEEEbib}
\bibliography{icme2019}
\end{document}